# Text classification dataset and analysis for Uzbek language

**Elmurod Kuriyozov[1,2], Ulugbek Salaev[1], Sanatbek Matlatipov[3], Gayrat Matlatipov[1]**

[1]Urgench State University, 14, Kh.Alimdjan str, Urgench city, 220100, Uzbekistan
{elmurod1202, ulugbek.salaev,gayrat}@urdu.uz

[2]Universidade da Coruña, CITIC, Grupo LYS, Depto. de Computación y Tecnologías de la Información, Facultade de Informática, Campus de Elviña, A Coruña 15071, Spain
e.kuriyozov@udc.es

National University of Uzbekistan named after Mirzo Ulugbek, 4 Universitet St, Tashkent, 100174, Uzbekistan
s.matlatipov@nuu.uz

**Abstract**

Text classification is an important task in Natural Language Processing (NLP), where the goal is to categorize text data into predefined classes. In this study, we analyze the dataset creation steps and evaluation techniques of multi-label news categorisation task as part of text classification. We first present a newly obtained dataset for Uzbek text classification, which was collected from 10 different news and press websites and covers 15 categories of news, press and law texts. We also present a comprehensive evaluation of different models, ranging from traditional bag-of-words models to deep learning architectures, on this newly created dataset. Our experiments show that the Recurrent Neural Network (RNN) and Convolutional Neural Network (CNN) based models outperform the rule-based models. The best performance is achieved by the BERTbek model, which is a transformer-based BERT model trained on the Uzbek corpus. Our findings provide a good baseline for further research in Uzbek text classification.

## 1. Introduction

Text classification is a critical task in the field of natural language processing (NLP), where the goal is to categorize a text document into predefined classes. This task is essential in many real-world applications such as sentiment analysis, spam detection, and topic modelling. With the massive amount of unstructured data generated daily, text classification provides a means to make sense of this data and derive meaningful insights.

In recent years, deep learning models have been widely used in text classification (Minaee et al., 2021), yielding excellent results. However, most research works in text classification have focused on high-resource languages such as English (Cruz & Cheng, 2020). There is a significant gap in text classification research for low-resource languages, Uzbek being no exception.

The primary objective of this work is to contribute to the NLP research community by addressing the text classification challenge for the Uzbek language, in the example of a multi-label news categorization task. We present a new Uzbek text classification dataset and evaluate the performance of various models on this dataset. The models range from traditional rule-based approaches, such as word and character n-grams-based support vector machine (SVM), to more advanced deep learning models, such as recurrent neural networks (RNN) and convolutional neural networks (CNN). We also analyse the dataset further by using transformer-based models, such as mBERT - a multilingual BERT model trained on more than 100 languages (Devlin et al., 2019), and BERTbek - a monolingual BERT language model trained on an Uzbek news corpus. Our experiment results indicate that neural-network-based models outperform the rule-based ones, and the BERTbek model achieves the best result with over 85% of the F1-score.

**Uzbek language.**
The Uzbek language is spoken by over 30 million people and is primarily used in Uzbekistan and surrounding Central Asian countries. It is a Turkic language that has been heavily influenced by both Russian, Arabic and Persian for geographic and historical reasons. As a low-resource language, there is limited research and resources available for Natural Language Processing (NLP) tasks in Uzbek, making the creation and utilization of NLP resources a crucial step towards promoting the digitalization of the Uzbek language. Despite this, Uzbek has a rich literary history and continues to be an important part of the cultural heritage of the Uzbek people.

Its official alphabet is Latin, and its grammar is close to other languages in the Turkic family, which differs vastly from the more commonly studied languages in NLP such as English and Chinese. This presents a challenge for NLP tasks in Uzbek, as models trained on those languages may not be effective in handling the nuances of Uzbek text. The development of NLP resources and models specifically for Uzbek can help advance research in the field and promote the use of technology in Uzbek-speaking communities[1].

The rest of the paper is organized as follows: We provide an overview of text classification and highlight some recent NLP works on Uzbek in Section 2. It is followed by the Methodology in section 3, where we describe the data collection and dataset creation process. In the Experiments section (Section 4), we describe the models used for evaluation. Moving on, Section 5 covers the results of the experiments and is followed by Section 6, where we discuss the effects and their implications. Finally, in the Conclusion and Future Work section (Section 7), we provide a conclusion of the work and outline future directions.

---

[1] More about the Uzbek language:
https://en.wikipedia.org/wiki/Uzbek_language

## 2. Related work

Text classification has been a fundamental problem in the field of Natural Language Processing (NLP) and has numerous applications in various domains such as sentiment analysis (Medhat et al., 2014), spam detection (Jindal & Liu, 2007), and categorization of news articles (Haruechaiyasak et al., 2008). With the advancement of machine learning techniques, the performance of text classification has improved dramatically in recent years. In the early days, traditional machine learning methods such as Support Vector Machines (SVM) (Joachims & others, 1999) and Naive Bayes (McCallum et al., 1998) were used for text classification. However, the growing size of text data and the increased complexity of the tasks led to the development of deep learning methods.

One of the major breakthroughs in text classification was the use of Convolutional Neural Networks (CNNs) for sentiment analysis by Kim (Kim & Lee, 2014). This work showed that the use of convolutional layers with different kernel sizes could effectively capture local and global information from texts. Recurrent Neural Networks (RNNs) have also been widely used for text classification tasks due to their ability to model sequential data. LSTMs, GRUs, and Bi-LSTMs have been popular variants of RNNs for text classification (Liu et al., 2016; Minaee et al., 2021). The use of attention mechanisms has further improved the performance of text classification tasks. The Transformer architecture introduced by Vaswani et al. (Vaswani et al., 2017) revolutionized the NLP field with its self-attention mechanism, and the BERT model (Devlin et al., 2018) based on the Transformer architecture has become a benchmark in various NLP tasks including text classification.

**NLP works on the Uzbek language.**

Despite the fact that Uzbek is considered a low-resource language, there have been some efforts to develop NLP resources and models for it. Some notable works include the creation of sentiment analysis datasets (Kuriyozov et al., 2022; Matlatipov et al., 2022), semantic evaluation datasets (Salaev et al., 2022b), and stopwords datasets (Madatov et al., 2022). NLP tools such as part-of-speech taggers (Sharipov et al., 2023), stemmers, and lemmatizers (Sharipov & Yuldashov, 2022) have also been developed to support NLP research and applications on Uzbek texts. However, further efforts are needed to improve the performance of NLP models on Uzbek texts.

Rabbimov and Kobilov (Rabbimov & Kobilov, 2020) focus on a similar task of multi-class text classification for texts written in Uzbek. The authors try to create a functional scheme of text classification and develop models using six different machine learning algorithms, including Support Vector Machines (SVM), Decision Tree Classifier (DTC), Random Forest (RF), Logistic Regression (LR) and Multinomial Naïve Bayes (MNB). The authors used the TF-IDF algorithm and word-level and character-level n-gram models as feature extraction methods and defined hyperparameters for text classification using 5-fold cross-validation. Through experiments conducted on a dataset developed from articles on ten categories from the Uzbek "Daryo" online news edition, the authors achieved a high accuracy of 86.88%. The only drawbacks of this paper are that the dataset is only limited to a single news source, hence working on a relatively small amount of data, the categories are also limited to ten classes, and the analysis is limited to machine learning techniques. We aim to fill these gaps in our current work by collecting more data, creating more text classes, as well as analysing the new dataset with deep learning models.

## 3. Methodology

In this section, we describe the steps of data collection in detail, as well as the efforts taken to clear the collected data, make some adjustments, and create the text classification dataset.

### 2.1. Data collection

Since text classification requires a labelled dataset for training and evaluating the models. For our research, we collected text data from 10 different Uzbek news websites, as well as press portals, including news articles and press releases. The websites were chosen to represent a diverse range of categories, such as politics, sports, entertainment, technology, etc. The data was collected using web scraping techniques, such as Scrapy framework for Python[2] and Beautiful Soup[3] preserving the source link, source category name, its title, and the main body. Each article was labelled with its corresponding category information. The collected dataset consisted of approximately 513K articles with more than 120M words in total, providing a large and diverse corpus for text classification. All the names of sources, a number of articles obtained from each source, as well as some information regarding the volume of the text are presented in Table 1.

### 2.2. Dataset creation

The dataset creation process involved several steps to ensure the quality and sustainability of the data for text classification. First, repetitive news and law decrees were removed to eliminate redundancy in the data. References to images, emojis, and URLs were also removed to ensure the data only contained text relevant to the classification task.

Additionally, some of the crawled texts in the dataset were written in the Cyrillic script. To address this, the texts were transliterated into the Latin script using the UzTransliterator tool (Salaev et al., 2022a).

Initially, there were more than 40 distinct categories when all the news texts were collected, but many of them were either synonymous or very close to one another, belonging to the same field. To ensure a better representation and a balanced distribution of the data, categories with identical or very close labels and some categories with a very small number of news articles were merged together. This helped to avoid the model getting confused over categories of very similar fields, as well as being biased towards certain categories with a larger number of samples.

---

[2] https://scrapy.org/

[3] https://pypi.org/project/beautifulsoup4/

| Category/Label | Source(s)* | # of Articles | % | # of Words | Avg. # of Words | Avg. # of Char-s |
|---|---|---|---|---|---|---|
| Local (Mahalliy) | 1, 3, 5 | 149312 | 29.1 | 34.7M | 232 | 1995 |
| World (Dunyo) | 1, 2, 3, 5 | 136732 | 26.7 | 21.1M | 155 | 1282 |
| Sport (Sport) | 1, 2, 3, 4, 5 | 59784 | 11.7 | 11.3M | 189 | 1512 |
| Society (Jamiyat) | 1, 2, 4, 5 | 55018 | 10.7 | 13.9M | 253 | 2114 |
| Law (Qonunchilik) | 6, 7 | 33089 | 6.5 | 27.0M | 815 | 7466 |
| Tech (Texnologiya) | 1, 2, 3, 5 | 17541 | 3.4 | 3.1M | 179 | 1467 |
| Culture (Madaniyat) | 2, 3 | 12798 | 2.5 | 2.9M | 226 | 1838 |
| Politics (Siyosat) | 1, 2, 4, 8 | 12247 | 2.4 | 3.4M | 279 | 2468 |
| Economics (Iqtisodiyot) | 1, 2, 4, 5 | 12165 | 2.4 | 3.1M | 257 | 2166 |
| Auto (Avto) | 3 | 6044 | 1.2 | 0.9M | 153 | 1273 |
| Health (Salomatlik) | 2, 3, 4 | 5086 | 1.0 | 1.3M | 257 | 2107 |
| Crime (Jinoyat) | 2 | 4200 | 0.8 | 0.8M | 181 | 1488 |
| Photo (Foto) | 1, 3 | 4037 | 0.8 | 0.6M | 150 | 1225 |
| Womens (Ayollar) | 3 | 2657 | 0.5 | 0.7M | 270 | 2156 |
| Culinary (Pazandachilik) | 3, 9 | 2040 | 0.4 | 0.1M | 62 | 498 |

*Notes: 1 - bugun.uz, 2 - darakchi.uz, 3 - daryo.uz, 4 - gazeta.uz, 5 - kun.uz, 6 - lex.uz, 7 - norma.uz, 8 - president.uz, 9 - zira.uz*

Table 1. Detailed information of the categories, names of their sources, percentage over the overall dataset, as well as the total and average number of words & characters per category.

All the above steps were taken to clean and pre-process the data and make it suitable for the text classification task. The final dataset consisted of a total of 512,750 news articles across 15 distinct categories, representing the Uzbek language as much as possible.

## 4. Experiments

For experiments on the newly created dataset, we randomly split the dataset with a 5:3:2 ratio for training, validation, and testing, respectively. During the splitting, we made sure that all the parts would have evenly distributed article categories.

In this study, we have carried out several experiments to evaluate the performance of different models on the Uzbek text classification task. The following models have been used for experiments:

- $LR_{Word-ngrams}$: Logistic regression with word-level n-grams (unigram and bi-gram bag-of-words models, with TF-IDF scores);
- $LR_{Character-ngrams}$: Logistic regression with character-level n-grams (bag-of-words model with up to 4-character n-grams);
- $LR_{Word+Char-ngrams}$: Logistic regression with word and character-level n-grams (concatenated word and character TF-IDF matrices);
- **RNN:** Recurrent neural network without pretrained word embeddings (bidirectional GRU with 100 hidden states, the output of the hidden layer is the concatenation of the average and max pooling of the hidden states);
- $RNN_{Word-embeddings}$: Recurrent neural networks with pretrained word embeddings (previous bidirectional GRU model with the SOTA 300-dimensional FastText word embeddings for Uzbek obtained from (Kuriyozov et al., 2020));
- **CNN:** Convolutional neural networks (multi-channel CNN with three parallel channels, kernel sizes of 2, 3 and 5; the output of the hidden layer is the concatenation of the max pooling of the three channels);
- **RNN + CNN:** RNN + CNN model (convolutional layer added on top of the GRU layer);
- **mBERT**: Multilingual BERT model, trained using more than a hundred languages, (including Uzbek) (Devlin et al., 2019);
- **BERTbek:** Monolingual BERT model trained on Uzbek news corpus[4].

We trained each model with the training dataset, fine-tuned using the evaluation dataset, and tested the model performance using the test dataset.

The rule-based models have been used as baselines to measure the performance of the neural network models. The *RNN* and *CNN* models were used to explore the ability of the recurrent and convolutional neural networks to capture the sequence information and the semantic representation of the Uzbek text data. Finally, the *BERT* model was used to evaluate the performance of the state-of-the-art language representation model in the Uzbek text classification task.

## 5. Results

In this section, we present the results of our experiments with the different models used for text classification on the Uzbek language dataset. We evaluated the performance of our models using several metrics including accuracy, F1-score, and precision. For each category in the dataset, the F1-scores of all experiment models and their mean scores are reported in Table 2.

Based on the model performance results, it can be concluded that the logistic regression models work best when both the word level and character level n-grams are considered (by concatenating their TF-IDF matrices).

---
[4] The BERTbek-news-big-cased model was used from https://huggingface.co/elmurod1202/BERTbek

| Models | F1 | Local | World | Sport | Society | Law | Tech | Culture | Politics | Economics | Auto | Health | Crime | Photo | Women | Culinary |
|---|---|---|---|---|---|---|---|---|---|---|---|---|---|---|---|---|
| $LR_{Word\text{-}ngram}$ | 73.6 | 89.8 | 86.5 | 79.2 | 62.3 | 76.1 | 63.4 | 66.3 | 77.1 | 74.5 | 80.7 | 69.2 | 72.2 | 68.5 | 61.2 | 77.1 |
| $LR_{Char\text{-}ngram}$ | 72.5 | 88.5 | 89.7 | 76.8 | 60.1 | 77.0 | 60.3 | 64.4 | 75.9 | 73.7 | 81.4 | 71.2 | 68.3 | 65.7 | 60.5 | 74.1 |
| $LR_{Word+Char\text{-}ngram}$ | 75.6 | 91.1 | 90.1 | 81.7 | 66.0 | 73.5 | 65.0 | 68.4 | 81.4 | 77.5 | 83.1 | 71.9 | 74.9 | 67.7 | 63.1 | 79.4 |
| $RNN$ | 79.0 | 91.5 | 92.4 | 86.1 | 64.9 | 82.7 | 66.0 | 71.6 | 84.1 | 79.7 | 88.7 | 79.2 | 77.2 | 70.5 | 67.8 | 82.5 |
| $RNN_{Word\text{-}emb.}$ | 80.4 | 93.6 | **93.0** | 88.1 | 66.8 | 81.6 | 66.9 | 73.4 | 82.9 | 82.5 | 89.1 | 82.5 | 80.5 | 73.7 | 66.9 | 83.9 |
| $CNN$ | 80.8 | 92.6 | 90.5 | 92.5 | 68.9 | 86.3 | 64.3 | 69.4 | 86.2 | 82.6 | 90.8 | 80.7 | 82.1 | 70.9 | 64.1 | 90.6 |
| $RNN + CNN$ | 83.3 | 94.0 | 92.3 | 94.1 | 72.4 | 84.6 | **68.4** | 74.0 | 86.7 | 86.1 | 92.1 | 83.7 | **85.7** | 75.0 | 69.5 | 91.0 |
| $mBERT$ | 83.4 | 92.1 | 91.2 | **93.5** | 74.7 | 89.5 | 67.6 | 76.8 | 89.4 | 86.6 | 91.4 | 86.5 | 83.5 | 71.8 | 67.3 | 89.5 |
| $BERTbek$ | **85.2** | **94.1** | 93.0 | 93.2 | **74.9** | **91.5** | 67.1 | **78.7** | **90.0** | **88.2** | **93.4** | **88.2** | 85.6 | **75.8** | **71.7** | **93.3** |

Table 2. Text classification evaluation results for all models. F1 scores per model and category and their mean values are reported, best scores overall and for each category are highlighted.

Neural network models, such as $RNN$ and $CNN$, perform better than rule-based models, and their performance is enhanced by adding specific knowledge of the language, such as pretrained word-embedding vectors. Among the transformer-based models, the monolingual $BERTbek$ model achieved the highest performance with an F1-score of 85.2%, compared to its multilingual counterpart (with 83.4% F1-score).

The results of our experiments demonstrate the effectiveness of deep learning models for text classification in the Uzbek language and provide a strong foundation for further research in this area.

## 6. Discussion

Analysing the performance results of the models over the newly obtained dataset, one can say that the text distribution of the news data over categories plays an important role, as the categories with significantly more data (such as *Local*, *World*, *Law*, etc.) achieve higher performance results, overall evaluation models, compared to other categories. The counter-wise situation is also true since some categories with very small amounts of data (such as *Women*, *Photo*, *Culture*, etc.) perform less overall.

Some categories with distinct keywords that are only used in their own field, such as *Sport* (most common keywords: sports names, and names of teams and players), *Auto* (most common keywords: car brands), as well as *Culinary* (most common keywords: names of ingredients, cooking terms), that can be easily predicted also reflect in the overall models' performance, showing high scores for those categories. Although the category *Tech* can be easily predicted like the previously-mentioned categories, it achieves the lowest performance scores in our case, due to the fact that the news data in that category look like other categories like *Auto* and *Photo*, making it hard for the models to predict the labels right.

Lastly, it can also be observed that the monolingual $BERTbek$ model outperforms the multilingual $mBERT$ model in many cases, due to the fact that the multilingual model includes a very small portion of texts in Uzbek. Only in the cases of predicting the labels for the *Tech* and *Sport* categories, $mBERT$ outperforms the $BERTbek$, which is caused by the fact that most of the key terms used in those texts are either named entities or international terms.

## 7. Conclusion and Future Work

In this paper, we aimed to tackle the task of text classification for the low-resource Uzbek language. Our contribution to the field includes a new dataset consisting of more than 512K labelled news texts with more than 120M words, spanned over 15 categories collected from 10 different news and press websites. The dataset was pre-processed to remove unwanted text, such as duplicates, references to images, emojis, and URLs, and transliterated from Cyrillic to Latin. In our experiments, we compared the performance of various models including rule-based models, deep learning models, as well as multilingual and monolingual transformer-based language models.

Our evaluation results showed that the BERT-based models outperform other models, while the monolingual BERT-based model achieved the highest score.

In conclusion, we have shown that deep learning models can effectively handle text classification tasks for the Uzbek language. In future work, we plan to improve the performance of the models by fine-tuning them on a larger dataset, and also to extend the study to other NLP tasks such as sentiment analysis, named entity recognition, and machine translation. Furthermore, we aim to develop open-source tools to make Uzbek NLP resources easily accessible to researchers and practitioners in the field.

## Data availability

The newly created Uzbek text classification dataset and the Python codes used for the evaluation of the models are publicly available at the project repository[5] as well as an open-access data platform[6].

This dataset will serve as a valuable resource for further NLP research on Uzbek language, and we hope it will stimulate further work in this area. By making the data and codes openly accessible, we aim to foster reproducibility and collaboration in the field.

---

[5] https://github.com/elmurod1202/TextClassification

[6] https://doi.org/10.5281/zenodo.7677430

## Acknowledgements

This research work was fully funded by the REP-25112021/113 - "UzUDT: Universal Dependencies Treebank and parser for natural language processing on the Uzbek language" subproject funded by The World Bank project "Modernizing Uzbekistan national innovation system" under the Ministry of Innovative Development of Uzbekistan.

## Declarations

The authors declare no conflict of interest. The founding sponsors had no role in the design of the study; in the collection, analysis, or interpretation of data; in the writing of the manuscript, and in the decision to publish the results.

## References

Cruz, J. C. B., & Cheng, C. (2020). Establishing baselines for text classification in low-resource languages. *ArXiv Preprint ArXiv:2005.02068*.

Devlin, J., Chang, M.-W., Lee, K., & Toutanova, K. (2018). Bert: Pre-training of deep bidirectional transformers for language understanding. *ArXiv Preprint ArXiv:1810.04805*.

Devlin, J., Chang, M.-W., Lee, K., & Toutanova, K. (2019). BERT: Pre-training of Deep Bidirectional Transformers for Language Understanding. *Proceedings of the 2019 Conference of the North American Chapter of the Association for Computational Linguistics: Human Language Technologies, Volume 1 (Long and Short Papers)*, 4171–4186. https://doi.org/10.18653/v1/N19-1423

Haruechaiyasak, C., Jitkrittum, W., Sangkeettrakarn, C., & Damrongrat, C. (2008). Implementing news article category browsing based on text categorization technique. *2008 IEEE/WIC/ACM International Conference on Web Intelligence and Intelligent Agent Technology*, *3*, 143–146.

Jindal, N., & Liu, B. (2007). Review spam detection. *Proceedings of the 16th International Conference on World Wide Web*, 1189–1190.

Joachims, T., & others. (1999). Transductive inference for text classification using support vector machines. *Icml*, *99*, 200–209.

Kim, J., & Lee, M. (2014). Robust lane detection based on convolutional neural network and random sample consensus. *Neural Information Processing: 21st International Conference, ICONIP 2014, Kuching, Malaysia, November 3-6, 2014. Proceedings, Part I 21*, 454–461.

Kuriyozov, E., Doval, Y., & Gomez-Rodriguez, C. (2020). Cross-Lingual Word Embeddings for Turkic Languages. *Proceedings of The 12th Language Resources and Evaluation Conference*, 4054–4062.

Kuriyozov, E., Matlatipov, S., Alonso, M. A., & Gómez-Rodr\'\iguez, C. (2022). Construction and evaluation of sentiment datasets for low-resource languages: The case of Uzbek. *Human Language Technology. Challenges for Computer Science and Linguistics: 9th Language and Technology Conference, LTC 2019, Poznan, Poland, May 17–19, 2019, Revised Selected Papers*, 232–243.

Liu, P., Qiu, X., & Huang, X. (2016). Recurrent neural network for text classification with multi-task learning. *ArXiv Preprint ArXiv:1605.05101*.

Madatov, K., Bekchanov, S., & Vičič, J. (2022). *Automatic detection of stop words for texts in the Uzbek language*.

Matlatipov, S., Rahimboeva, H., Rajabov, J., & Kuriyozov, E. (2022). Uzbek Sentiment Analysis Based on Local Restaurant Reviews. *CEUR Workshop Proceedings*, *3315*, 126–136. www.scopus.com

McCallum, A., Nigam, K., & others. (1998). A comparison of event models for naive bayes text classification. *AAAI-98 Workshop on Learning for Text Categorization*, *752*(1), 41–48.

Medhat, W., Hassan, A., & Korashy, H. (2014). Sentiment analysis algorithms and applications: A survey. *Ain Shams Engineering Journal*, *5*(4), 1093–1113.

Minaee, S., Kalchbrenner, N., Cambria, E., Nikzad, N., Chenaghlu, M., & Gao, J. (2021). Deep learning–based text classification: a comprehensive review. *ACM Computing Surveys (CSUR)*, *54*(3), 1–40.

Rabbimov, I. M., & Kobilov, S. S. (2020). Multi-class text classification of uzbek news articles using machine learning. *Journal of Physics: Conference Series*, *1546*(1), 12097.

Salaev, U., Kuriyozov, E., & Gómez-Rodríguez, C. (2022a). A Machine Transliteration Tool Between Uzbek Alphabets. *CEUR Workshop Proceedings*, *3315*.

Salaev, U., Kuriyozov, E., & Gómez-Rodríguez, C. (2022b). SimRelUz: Similarity and Relatedness scores as a Semantic Evaluation Dataset for Uzbek Language. *1st Annual Meeting of the ELRA/ISCA Special Interest Group on Under-Resourced Languages, SIGUL 2022 - Held in Conjunction with the International Conference on Language Resources and Evaluation, LREC 2022 - Proceedings*.

Sharipov, M., Kuriyozov, E., Yuldashev, O., & Sobirov, O. (2023). UzbekTagger: The rule-based POS tagger for Uzbek language. *ArXiv Preprint ArXiv:2301.12711*.

Sharipov, M., & Yuldashov, O. (2022). UzbekStemmer: Development of a Rule-Based Stemming Algorithm for Uzbek Language. *ArXiv Preprint ArXiv:2210.16011*.

Vaswani, A., Shazeer, N., Parmar, N., Uszkoreit, J., Jones, L., Gomez, A. N., Kaiser, Ł., & Polosukhin, I. (2017). Attention is all you need. *Advances in Neural Information Processing Systems*, *30*.